\documentclass[10pt,twocolumn,letterpaper]{article}

\usepackage{authblk} % added by Matze
\usepackage{longtable}
\usepackage{xcolor}
\usepackage{appendix}

\definecolor{brandeisblue}{rgb}{0.0, 0.44, 1.0}

\usepackage[pagenumbers]{cvpr} % To force page numbers, e.g. for an arXiv version

\usepackage{graphicx}
\usepackage{amsmath}
\usepackage{amssymb}
\usepackage{booktabs}
\usepackage[accsupp]{axessibility}  % Improves PDF readability for those with disabilities.

\usepackage[pagebackref,breaklinks,colorlinks,urlcolor=brandeisblue]{hyperref}

\usepackage[capitalize]{cleveref}
\crefname{section}{Sec.}{Secs.}
\Crefname{section}{Section}{Sections}
\Crefname{table}{Table}{Tables}
\crefname{table}{Tab.}{Tabs.}

\newcounter{appendix}

\newcommand{\appendixtitle}{}
\newcommand{\appendixtitleFull}{%
  \space\Alph{appendix}.\quad\appendixtitle}
\AtBeginEnvironment{appendices}{%
  \captionsetup{labelsep=period}
  \counterwithin{figure}{appendix}
  \counterwithin{table}{appendix}
  \counterwithin{equation}{appendix}
  \counterwithin{section}{appendix}  
  \preto{\section}{%
    \clearpage
    \refstepcounter{appendix}
    \phantomsection 
    \addcontentsline{toc}{section}{\appendixtitleFull}%
  }
}

\def\cvprPaperID{9999} % *** Enter the CVPR Paper ID here
\def\confName{CVPR}
\def\confYear{2023}

\begin{document}

\title{Why is the winner the best?}

\author[1, 2]{M. Eisenmann}
\author[1, 2, 3]{A. Reinke}
\author[4]{V. Weru}
\author[1, 2]{M. D. Tizabi}
\author[5, 2]{F. Isensee}
\author[1]{T. J. Adler}
\author[6]{S. Ali}
\author[7, 8]{\authorcr V. Andrearczyk}
\author[9]{M. Aubreville}
\author[10, 11, 12]{U. Baid}
% ...

\affil[1]{\small Division of Intelligent Medical Systems, German Cancer Research Center (DKFZ), Heidelberg, Germany}
\affil[2]{Helmholtz Imaging, German Cancer Research Center (DKFZ), Heidelberg, Germany
\authorcr Full affiliations given in App.~\ref{app:affiliation_list}.}

\maketitle

%%%%%%%%% ABSTRACT
\begin{abstract}
International benchmarking competitions have become fundamental for the comparative performance assessment of image analysis methods. However, little attention has been given to investigating what can be learnt from these competitions. Do they really generate scientific progress? What are common and successful participation strategies? What makes a solution superior to a competing method? To address this gap in the literature, we performed a multi-center study with all 80 competitions that were conducted in the scope of IEEE ISBI 2021 and MICCAI 2021. Statistical analyses performed based on comprehensive descriptions of the submitted algorithms linked to their rank as well as the underlying participation strategies revealed common characteristics of winning solutions. These typically include the use of multi-task learning (63\%) and/or multi-stage pipelines (61\%), and a focus on augmentation~(100\%), image preprocessing (97\%), data curation (79\%), and postprocessing (66\%). The ``typical'' lead of a winning team is a computer scientist with a doctoral degree, five years of experience in biomedical image analysis, and four years of experience in deep learning. Two core general development strategies stood out for highly-ranked teams: the reflection of the metrics in the method design and the focus on analyzing and handling failure cases. According to the organizers, 43\% of the winning algorithms exceeded the state of the art but only 11\% completely solved the respective domain problem. The insights of our study could help researchers (1) improve algorithm development strategies when approaching new problems, and (2) focus on open research questions revealed by this work.
\end{abstract}

\begin{figure}
  \centering
  \includegraphics[width=1.0\linewidth]
                   {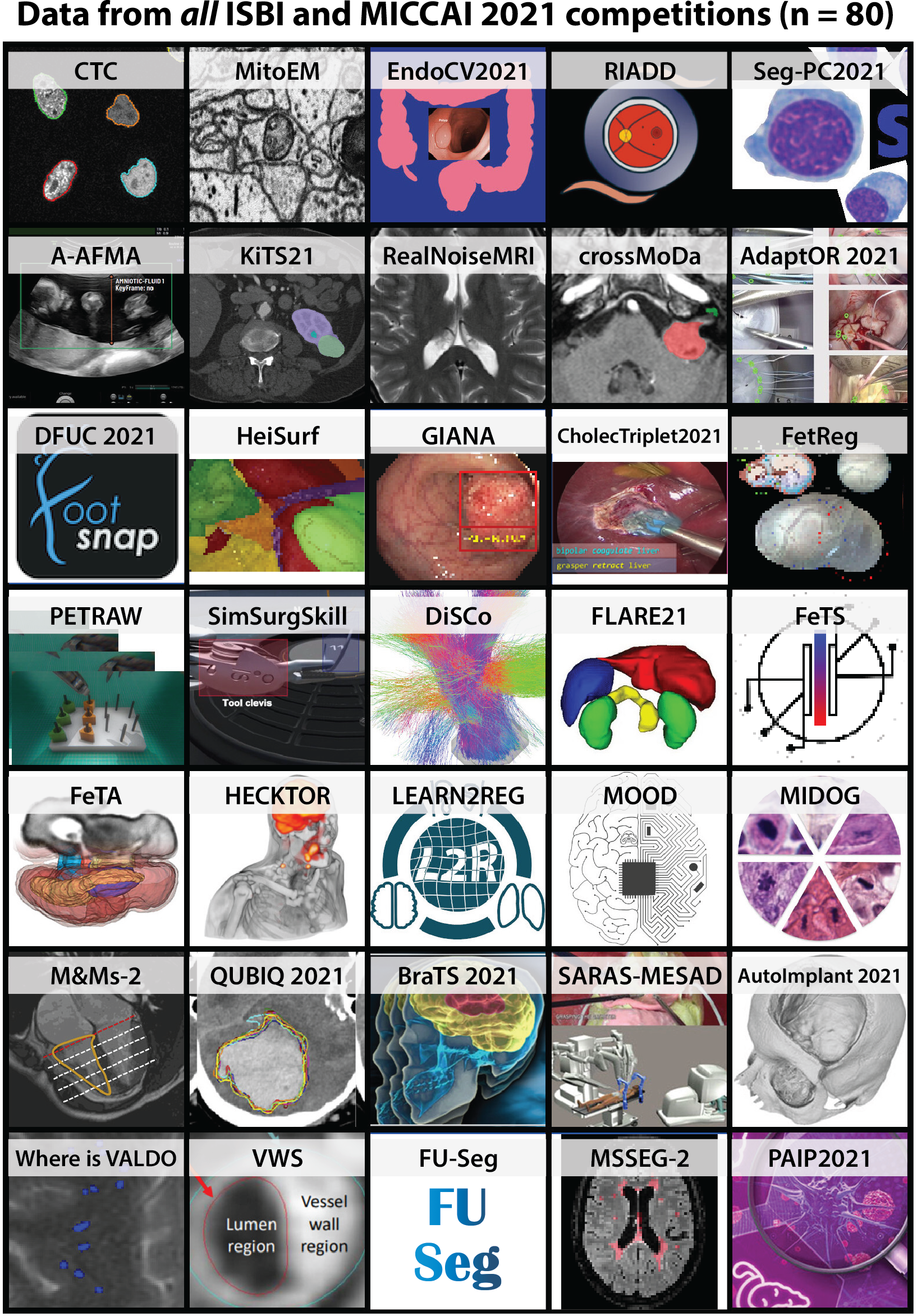}
  \caption{Overview of the IEEE ISBI 2021 and MICCAI 2021 challenges. Under the umbrella of 35 challenges (each represented by a teaser image and acronym), a total of 80 competitions with dedicated leaderboards were organized, as detailed in App.~\ref{app:overview}. We used data from participants, organizers, and winners to address the key research questions of this contribution: \textit{(RQ1) What is common practice in challenge participation?}, \textit{(RQ2) Do current competitions generate scientific progress?}, and \textit{(RQ3) Which strategies characterize challenge winners?}}
  \label{fig:teaser}
\end{figure}

%%%%%%%%% BODY TEXT
\section{Introduction}
\label{sec:intro}

Validation of biomedical image analysis algorithms is typically conducted through so-called challenges -- large international benchmarking competitions that compare algorithm performance on datasets addressing specific problems. Recent years have not only seen an increase in the complexity of the machine learning (ML) models used to solve the tasks, but also a substantial increase in the scientific impact of challenges, with results often being published in prestigious journals (e.g., \cite{sage2015quantitative, chenouard2014objective, menze2014multimodal, ulman2017objective, maier2017challenge}), and winners receiving tremendous attention in terms of citations and (sometimes) high monetary compensation~\cite{noauthor_kaggle_nodate}. However, despite this impact, little effort has so far been invested in investigating what can be learnt from a challenge. Firstly, we identified a notable gap in literature regarding insights into current common practices in challenges as well as studies that critically analyze whether challenges actually generate scientific progress. Secondly, while recent work has addressed the problem of deriving meaningful conclusions from challenges~\cite{wiesenfarth_methods_2021, maier-hein_why_2018}, it still remains largely unclear what makes winners the best and hence what constitutes a good strategy for approaching a new challenge or problem. The specific questions are manifold, e.g., \textit{Which specific training paradigms are used in current winning solutions?}, \textit{What are the most successful strategies for achieving generalization?}, \textit{Is it beneficial to involve domain experts or to work in a large team?}. While ablation studies on the effects of ML model component removal could be used to address some questions, they suffer from the major drawback of only providing insights into submitted solutions, but not into underlying strategies. Furthermore, they typically only allow for investigating few aspects of a solution, and come at the cost of a substantial carbon footprint.

To overcome these issues, we chose an approach that allowed us to systematically assess all of the aforementioned questions related to biomedical image analysis competitions within one cohesive study. To this end, members of the Helmholtz Imaging Incubator (HI) and of the Medical Image Computing and Computer Assisted Intervention (MICCAI) Special Interest Group on biomedical image analysis challenges designed a series of comprehensive international surveys that were issued to participants, organizers, and winners of competitions conducted within the IEEE International Symposium on Biomedical Imaging (ISBI) 2021 and the International Conference on MICCAI 2021. By collaborating with the organizers of all 80 competitions (100\%, see overview in App.~\ref{app:overview}), we were able to link algorithmic design decisions and challenge participation strategies to the outcome captured in rankings. Based on the study data, we explicitly addressed three research questions: \textit{(RQ1) What is common practice in challenge participation?}, \textit{(RQ2) Do current competitions generate scientific progress?}, and \textit{(RQ3) Which strategies characterize challenge winners?}

%-------------------------------------------------------------------------
\section{Methods}
\label{sec:methods}

According to the Biomedical Image Analysis ChallengeS (BIAS) Enhancing the QUAlity and Transparency Of health Research (EQUATOR) guideline on biomedical challenges~\cite{maier-hein_bias_2020}, a biomedical image analysis challenge is defined as an ``[...] open competition on a specific scientific problem in the field of biomedical image analysis. A challenge may encompass multiple competitions related to multiple \emph{tasks}, whose participating teams may differ and for which separate rankings/leaderboards/results are generated.''. As the term \emph{challenge task} is uncommon in the ML community, we will use the term \emph{competition} instead. The term \emph{challenge} will be reserved for the collection of tasks that are performed under the umbrella of one dedicated organization, represented by an acronym (Fig.~\ref{fig:teaser}). For our analyses, we targeted three main groups that are relevant in the context of challenges, namely (1) challenge participants, (2) challenge organizers, and (3) challenge winners. The following sections present the methodology developed to address the corresponding research questions RQ1-RQ3.

%-------------------------------------------------------------------------
\subsection{RQ1: What is common practice in challenge participation?}
\label{sec:methods:common_practice}

To investigate current common practice in biomedical image analysis challenge participation, we designed a survey that was addressed to challenge participants and structured in five parts covering: (1) general information on the team and the tackled task(s), (2) information on expertise and environment, (3) strategy for the challenge, (4) algorithm characteristics, and (5) miscellaneous information (details provided in Sec.~\ref{sec:results}).

The organizers of all IEEE ISBI 2021 challenges (30 competitions across 6 challenges \cite{noauthor_ctc_nodate, noauthor_mitoem_nodate, ali2022assessing, noauthor_riadd_nodate, noauthor_segpc-2021_nodate, noauthor_a-afma_nodate}), and all MICCAI 2021 challenges (50 competitions across 29 challenges \cite{nicholas_heller_2020_4674397, melanie_ganz_2021_4572640, reuben_dorent_2021_4573119, sandy_engelhardt_2021_4646979, moi_hoon_yap_2020_4646982, stefanie_speidel_2021_4572973, gabriel_girard_2021_4733450, jun_ma_2021_4596561, spyridon_bakas_2021_4573128, kelly_payette_2021_4573144, vincent_andrearczyk_2021_4573155, mattias_heinrich_2021_4573968, jens_petersen_2021_4573948, marc_aubreville_2021_4573978, carlos_martin_isla_2021_4573984, bjoern_menze_2021_4575204, spyridon_bakas_2021_4575162, fabio_cuzzolin_2021_4575197, jianning_li_2021_4577269, carole_sudre_2020_4600654, chun_yuan_2021_4575301, chuanbo_wang_2021_4575314, frederic_cervenansky_2021_4575409, jinwook_choi_2021_4575424}) were invited to participate in the initiative and to bring us into contact with participants (if allowed by the challenge privacy policy) or distribute the survey link to them. We created an individual survey website for each challenge to be able to accommodate the individual challenge submission deadline. To avoid bias in survey responses, participants were asked to complete the survey before knowing their position in the final ranking. Out of a maximum of 168 questions, the survey only showed questions that were relevant to the specific situation. The responses and feedback from the IEEE ISBI 2021 respondents were used to refine the survey for MICCAI 2021, and are thus not included in the results presented in \cref{sec:results:common_practice}.

Where organizers were allowed to share the contact details of the participants (20 challenges), the survey was conducted in closed-access mode, meaning that the participants received individual links to the survey and, where necessary, reminders. Fifteen surveys were conducted in open-access mode, meaning that the organizers were tasked with sharing the link to the respective survey and sending reminders. In these cases, we were not informed about the number of challenge participants and could not relate the number of responses to the total number.

%-------------------------------------------------------------------------
\subsection{RQ2: Do current competitions generate scientific progress?}
\label{sec:methods:scientific_progress}

The focus of the organizer survey was on the findings of the respective competition, particularly regarding whether scientific progress was made and, if yes, in which areas it was achieved and which open questions remain. To better put the respective competition into context, we also acquired general information on the associated competition(s).

%-------------------------------------------------------------------------
\subsection{RQ3: Which strategies characterize challenge winners?}
\label{sec:methods:winner_profile}

The complexity of state-of-the-art neural network-based approaches, involving numerous and interdependent design parameters, comes with the risk of attributing the success in a competition to the wrong component of a system. To approach the question \textit{Why is the winner the best?}, we linked the survey results of \cref{sec:methods:common_practice} to the final outcome of the competition and subsequently applied mixed model analyses. Given the large number of parameters relative to the number of competitions, we were aware that differences in parameters might not achieve statistical significance. In a second step, we therefore explicitly asked challenge winners for successful algorithm design choices and strategies in an additional survey.

%\subsubsection{Mixed model analysis}
%\label{sec:methods:winner_profile:mixed_model}

\textbf{Mixed model analysis} To compensate for the hierarchical data structure resulting from clusters corresponding to specific competitions, a logistic mixed model was used. In a first step, a univariable analysis was performed, i.e., the effect of each variable on the ranking was investigated separately. To further account for potential interdependencies between variables, two multivariable analyses were added. In the first analysis, the goal was to investigate the strategies influencing the probability of being the winner, while the second analysis focused on evaluating the strategies influencing the probability of being ranked among the best 30\%. For both analyses, a logistic mixed model was implemented. The winning strategies were included as fixed effects while the challenge identifier was included as a random effect. Additionally, some of the strategies were allowed to vary across challenges, specifically the total training time in computation hours, time spent on analyzing data and annotations, and time spent on analysis of failure cases. Variables with highly varying magnitudes were scaled before fitting the model. Statistical analysis was done in R Statistical Software~\cite{software_r} (v4.0.3, package: lme4~\cite{package_lme4}). 

\textbf{Survey on winning strategies} The survey of competition winners consisted of three main parts targeting the design decisions related to the winning submission, general recommended strategies for winning a competition, and the profile of a winner, respectively.

In the first part, we asked the winners about the importance of various design decisions for their submitted method. These comprised design decisions related to (1) the training paradigm, such as the usage of multi-task learning or semi-supervised learning, (2) network details, such as the choice of loss function(s), (3) model initialization, specifically pretraining, (4) data usage, covering aspects like data curation, augmentation, data splitting, and sampling, (5) hyperparameters, (6) ensembling, (7) postprocessing, and (8) metrics (see Fig.~\ref{fig:design_decisions}). For each of these design decisions, winners specified their method (e.g., whether they performed pretraining and, if so, based on which data) and rated the importance of this design choice for winning the challenge. We further explicitly asked what distinguished the winning solution from competing solutions and what were key factors for success.

The second part of the survey investigated general successful strategies (independent of the specific challenge). To this end, several authors of this paper who had already won multiple challenges compiled the list of strategies (Fig.~\ref{fig:winning_strategies}). The winners were asked to rate the importance of each strategy and further complement the list.

Finally, the third part of the survey covered questions on the profile of a challenge winner (Fig.~\ref{fig:key_insights_organizers}). This was particularly relevant for those winners that had not taken part in the original survey of \cref{sec:methods:common_practice}.

%-------------------------------------------------------------------------
\section{Results}
\label{sec:results}

Based on the positive responses of all organizers from all IEEE ISBI 2021 (n = 30) and MICCAI 2021 (n = 50) competitions, a total of 80 competitions conducted across 35 challenges were included in this study (Fig.~\ref{fig:teaser}). These covered a wide range of problems related to semantic segmentation, instance segmentation, image-level classification, tracking, object detection, registration, and pipeline evaluation.

%-------------------------------------------------------------------------
\subsection{Common practice in challenge participation}
\label{sec:results:common_practice}

A median (min/max) of 72\% (11\%/100\%) of the challenge participants took part in the survey, according to the closed-access surveys. Overall, we received 292 completed survey forms, of which 249 met our inclusion criteria (i.e., second version of the survey refined for MICCAI 2021, survey completed by a lead developer, no duplicate responses from the same team). Detailed responses to all aspects of the survey (including interquartile ranges (IQR) and min/max values of all parameters) are provided in a white paper \cite{eisenmann_biomedical_2022}. This section summarizes a selection of answers. The profile of a winner is depicted in Fig.~\ref{fig:winner_profile}.

\textbf{Infrastructure and strategies} Knowledge exchange was the most important incentive for participation (mentioned by 70\%; respondents were allowed to pick multiple answers), followed by the possibility to compare their own method to others (65\%), having access to data (52\%), being part of an upcoming challenge publication (50\%), and winning a challenge (42\%). The awards/prize money was important to only 16\% of the respondents. Regarding the computing infrastructure, only 25\% of all respondents thought that their infrastructure was a bottleneck. The vast majority of respondents used a Graphics Processing Unit (GPU) cluster. The total training time of all models trained during method development including failure models was estimated to be a median of 267 GPU hours, while the training time of the final submission was estimated to be a median of 24 GPU hours. %The most popular frameworks for method implementation, analyzing data, and analyzing annotations/reference data were PyTorch (76\%), NumPy (37\%), and NumPy (27\%), respectively.
The most popular frameworks were PyTorch for method implementation (76\%), NumPy for analyzing data (37\%), and NumPy for analyzing annotations/reference data (27\%).

The most common approach to development (42\%) consisted of going through related literature and building upon/modifying existing work. The majority (51\%) estimated the edited lines of code of the final solution to be in the order of magnitude of $10^3$. A median of 80 working hours was spent on method development in total. The respondents reported more human-driven decisions (median~of~60\%), e.g., parameter setting based on expertise, than empirical decisions (median~of~40\%), e.g., automated hyperparameter tuning via grid search. 94\% of the respondents used a deep learning-based approach. For those approaches, most time (up to three picks allowed) was spent on selecting one or multiple existing architectures that best matched the task (45\%), configuring the data augmentation (33\%), configuring the template architecture (e.g., How deep? How many stages/pooling layers?) (28\%), exploring existing loss functions (25\%), and ensembling (22\%).

The survey revealed that almost one third of the respondents did not have enough time for development. A majority thereof (65\%) felt that more time in the scale of weeks would have been beneficial (months: 18\%, days: 14\%).

\textbf{Algorithm characteristics} Among the deep learning-based approaches, only 9\% actively used additional data, i.e., data not provided for the respective challenge, in their final solution (note that this does not include the usage of already pretrained models). One reason may be that some challenges (24\%) explicitly do not allow the usage of external data. Of those that did leverage external data, the majority used public biomedical data for the same type of task (40\%), private biomedical data for the same type of task (25\%), or public biomedical data for a different type of task (15\%). Non-biomedical data was only used in 5\% of the cases. If additional data was used, it was used for pretraining (55\%) and/or co-training (50\%).

Data augmentation was applied by 85\% of the respondents. The most common augmentations were random horizontal flip (77\%), rotation (74\%), random vertical flip (62\%), contrast (49\%), scale (48\%), crop (44\%), resize crop (35\%), noise (34\%), elastic deformation (26\%), color jitter (19\%), and shearing (15\%). 43\% of the respondents reported that the data samples were too large to be processed at once (e.g., due to GPU memory constraints). This issue was mainly solved by patch-based training (cropping) (69\%), downsampling to a lower resolution (37\%), and/or solving 3D analysis tasks as a series of 2D analysis tasks (per z-slice approach) with postprocessing (18\%). The most common loss functions were Cross-Entropy (CE) Loss (39\%), combined CE and Dice Loss (32\%), and Dice Loss (26\%). 29\% of the respondents used early stopping, 12\% used warmup. Internal evaluation via a single train:val(:test) split was performed by more than half of the respondents (52\%). K-fold cross-validation on the training set was performed by 37\%. 6\% did not perform any internal evaluation. 48\% of the respondents applied postprocessing steps.

The final solution of 50\% of the respondents was a single model trained on all available data. An ensemble of multiple identical models, each trained on the full training set but with a different initialization (random seed), was proposed by 6\%. 21\% proposed an ensemble of multiple identical models, each trained on a randomly drawn subset of the training set (regardless of whether the same seed was used or not). 9\% reported having ensembled multiple different models and trained each on the whole training set (different seeds). 8\% ensembled multiple different models, each trained on a randomly drawn subset of the training set (regardless of whether the same seed was used or not). If multiple models were used, the final solution was composed of a median of 5 models.

%-------------------------------------------------------------------------
\subsection{Key insights related to scientific progress generated by challenges}
\label{sec:results:scientific_progress}

\begin{figure}
  \centering
  \includegraphics[width=1.0\linewidth]
                   {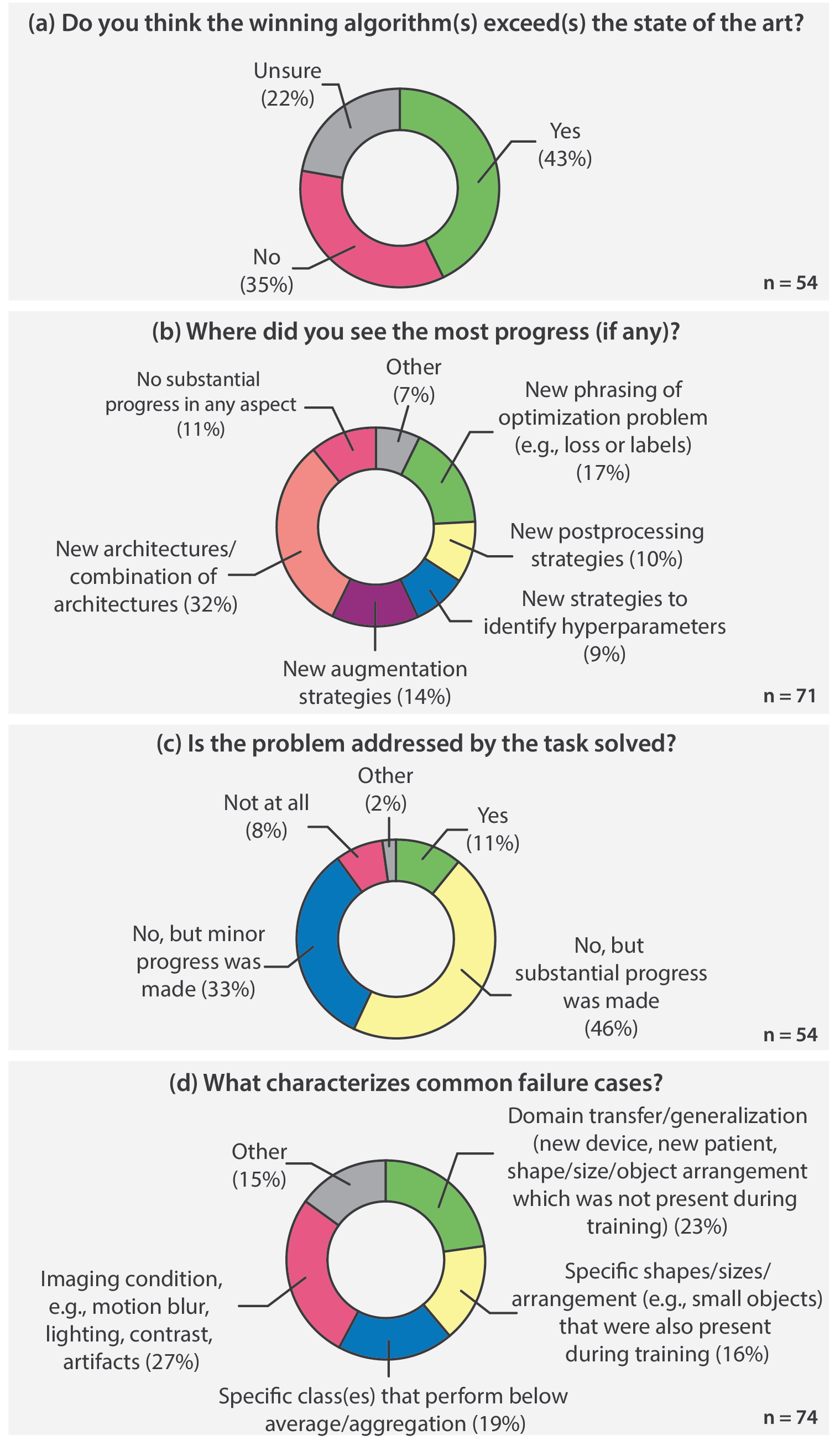}
  \caption{Key insights provided by the organizers of IEEE ISBI 2021 and MICCAI 2021 challenges.}
  \label{fig:key_insights_organizers}
\end{figure}

According to the responses of challenge organizers~(n~=~54), 43\% of the winning algorithms exceeded the state of the art (Fig.~\ref{fig:key_insights_organizers}). While substantial (47\%) or minor (32\%) progress was made in most competitions, the underlying problem was regarded as solved in only 11\% of the competitions. Most progress was seen in new architectures/combination of architectures (32\%), the phrasing of the optimization problem (e.g., new losses) (17\%,) and new augmentation strategies (14\%). Failure cases were mainly attributed to specific imaging conditions (e.g., image blur) (27\%), generalization issues (23\%), and specific classes that perform particularly poorly~(19\%).

According to the responses from several organizers, the trend of simple algorithms (e.g., U-Net~\cite{falk_u-net_2019}/nnU-Net~\cite{isensee_nnu-net_2021}) outperforming complex ones continued. As a prominent feature in 2021, many competitions provided additional information that is not usually available, such as the identifier of the hospital for domain generalization, multiple expert segmentations to represent label uncertainty, or k-space data in reconstruction problems. However, the participants were not able to leverage the additional data for better performance. The same holds true for temporal data in video analysis, although organizers hypothesize that frame-based analysis is not sufficient.

Several organizers also reported a lack of heterogeneity in methods. Often, submitted methods performed similarly (e.g., differing only in the fourth decimal digit in normalized scores). On a positive note, some competitions that had been run for multiple years observed a drastic improvement compared to previous years, sometimes even surpassing human performance. Regarding computational aspects, in one case the winning method surpassed the existing state-of-the-art method, achieving a 19 times faster inference speed and reduction of the GPU memory consumption by 60\% while yielding comparable accuracy.

According to our study, generalization remains a major issue. One challenge, which mimicked ``in-the-wild'' deployment, found that models failed to generalize in 3 out of 21 testing institutions. Similarly, performance in rare classes was reported as a core issue in several competitions. This is a problem of high clinical relevance as diseases often correspond to a rare class. A related problem is the fact that the detection of multiple conditions in a multi-label setting still remains challenging. Finally, some organizers reported the failure of metrics to reflect the biomedical domain interest. Along these lines, pixel-level performance was sometimes reported to be substantial while instance-/case-level performance, which is typically biomedically more relevant, was not improved substantially.

Cheating was observed in 4\% of the cases. It was related to an excessive number of submissions of similar methods with different user accounts or the attempt to retrieve the test set from the submission platform. In these cases, participants were excluded from the competition, the rankings and/or the publication.

\begin{figure*}
  \centering
  \includegraphics[width=1.0\linewidth]
                   {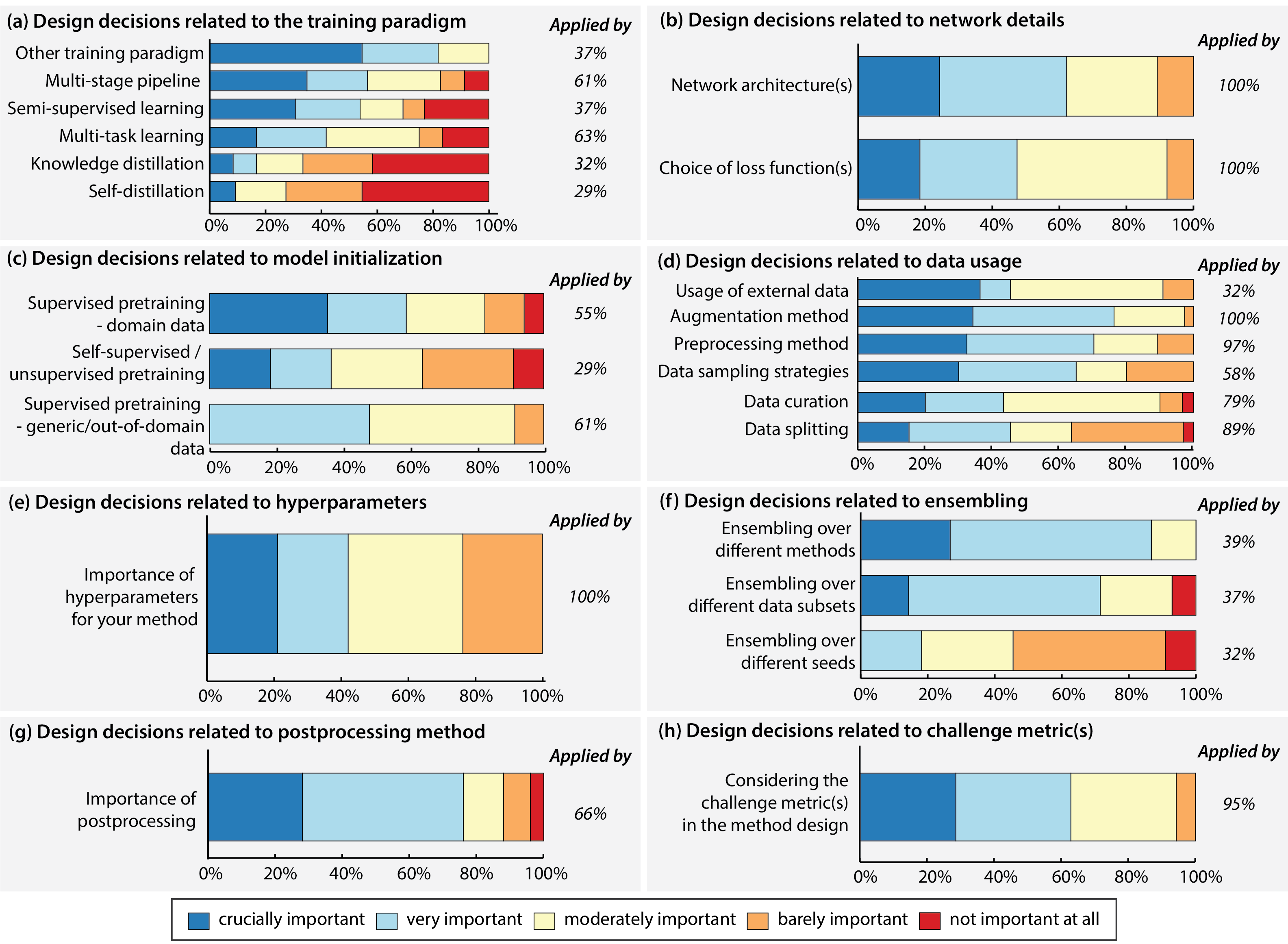}
  \caption{Importance of design decisions for the neural network-based winning submission of the respective IEEE ISBI 2021 and MICCAI 2021 competition rated by the (team) lead and ordered by percentage of highest vote (crucially important: dark blue). Voting was only conducted among those who used the respective design. “Applied by” indicates the percentage of respondents using the respective design.}
  \label{fig:design_decisions}
\end{figure*}

\begin{figure}
  \centering
  \includegraphics[width=1.0\linewidth]
                   {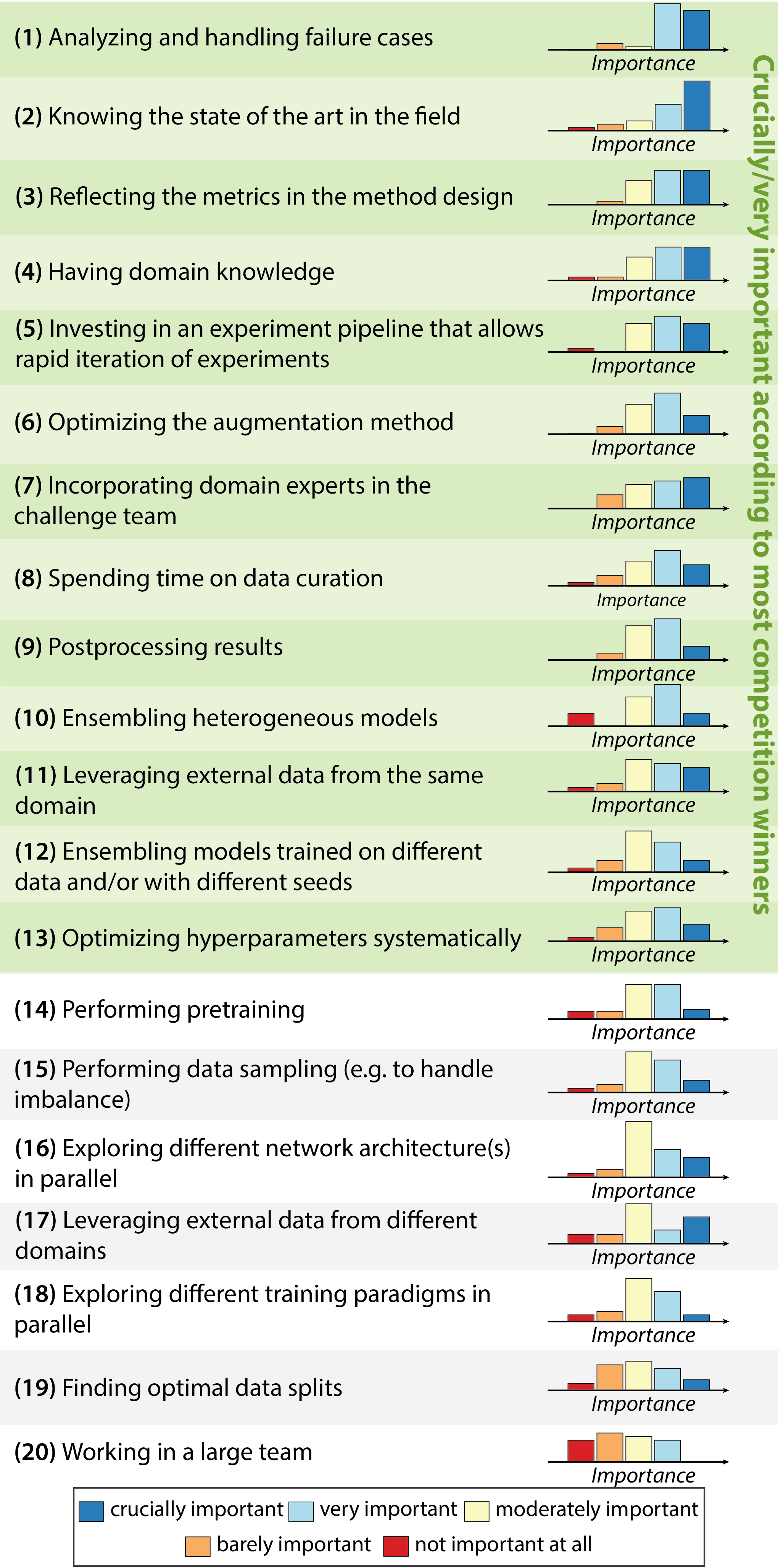}
  \caption{Strategies for winning a challenge according to winners of IEEE ISBI 2021 and MICCAI 2021 competitions, ordered by the sum of the ``crucially important'' (dark blue) and ``very important'' (light blue) categories. The distribution of importance (from left to right: not important at all, barely important, moderately important, very important, crucially important) is depicted for each strategy.}
  \label{fig:winning_strategies}
\end{figure}

%-------------------------------------------------------------------------
\subsection{Key insights related to winning strategies}
\label{sec:results:winner_profile}

When comparing winners to other participants, several differences stood out. Firstly, winners were more determined to win a challenge (64\% vs. 40\%). The majority of winning lead developers have a doctorate degree (41\%) while the majority of non-winning lead developers have a master's degree (47\%) as their highest degree. Furthermore, while only 66\% of other participants felt that there was enough development time, 86\% of the winners agreed with this statement. Winners spent 120~hours (e.g., on method development, analyzing data and annotations) before deciding to submit, compared to 56~hours for other participants, and decided to submit a week earlier (3 vs.~2 weeks prior to submission). Notably, winners spent twice as much time on failure analysis (10\% of median working hours dedicated to method development vs. 5\%). Compared to non-winners, winners used ensembling based on random seeds, data splits, and heterogeneous models (see Fig.~\ref{fig:design_decisions}(f)) 5.6~times, 1.7 times, and 2.5 times as much.

According to univariable mixed model analysis, eight parameters were found to provide statistically significant differences between winners and non-winners ($p~<~0.05$): (1) Number of team members who were developers/engineers, (2) time invested before planning to submit results, (3) time spent in data preprocessing/augmentation, (4) use of professionally managed GPU cluster, (5) approach used for method development, (6) architecture type, (7) taking metrics used to evaluate the challenge into account while searching for hyperparameters, and (8) augmentations used. Note, however, that when multiple independent tests are performed, 5\% can be expected to be identified as significant purely by chance when testing at 5\% significance level. Correcting for this so-called multiplicity of testing, we did not obtain statistically significant differences. Multivariable model analysis based on a selection of variables identified by image analysis experts revealed the willingness to win the challenge as the only parameter with $p~<~0.05$ when comparing winners to non-winners (64\% vs. 40\%). Analogously, the parameter of taking metrics used to evaluate the challenge into account while searching for hyperparameters was identified in the best 30\% vs. the rest analysis. It is worth mentioning in this context that despite the high response rate of 72\%, the number of winners covered by the survey presented in \cref{sec:methods:common_practice} was only 22. The resulting low power of identifying important contributors to winning challenges may well be the reason for the absence of statistical significance. We therefore additionally asked competition winners after the results announcement for key design decisions and strategies. The responses (n = 38) cover 67\% and 62\% of the IEEE ISBI 2021 and MICCAI 2021 challenges respectively, and are summarized in Fig.~\ref{fig:design_decisions} and Fig.~\ref{fig:winning_strategies}.

As detailed in Fig.~\ref{fig:design_decisions}, the most applied training pipelines were multi-task designs (63\%) and multi-stage pipelines (61\%). If multi-stage pipelines were applied, the importance of this strategy for winning the challenge was rated crucial. Pretraining was mainly performed in a supervised fashion using in-domain data (55\%) or generic data (e.g., ImageNet) (61\%). The usage of in-domain data, however, was found to be much more important. As mentioned above, it should be noted that many competitions do not allow for the usage of external data (24\% according to the survey presented in \cref{sec:methods:scientific_progress}). The most commonly applied design decisions related to data usage were preprocessing (97\%), augmentation (100\%), data splitting (beyond the splits provided by the competition, e.g., for cross-validation) (89\%), data curation (e.g., cleaning of annotations) (79\%), and data sampling (58\%). One aspect that stood out when asking winners for key factors for success (free text) was the setting up of a good internal validation strategy, including the careful selection of a baseline model and appropriate validation tests.

With respect to general strategies (Fig.~\ref{fig:winning_strategies}), the strategies of analyzing and handling failure cases, knowing the state of the art, and reflecting the metrics in the method design were rated most highly. Further recommended strategies in free-text answers were heterogeneous and comprise (1) inclusion of non-deep learning approaches in a model ensemble, (2) explicit determination of a time management strategy, (3) test-time augmentation, and (4) preferring matured architectures over brand-new hyped machine learning methods.

%-------------------------------------------------------------------------
\section{Discussion}
\label{sec:discussion}

The presented study represents, to the best of our knowledge, the first systematic and large-scale examination of biomedical image analysis competitions with a focus on what the scientific community can learn from them. Based on comprehensive surveys and statistical analyses for a total of 80 competitions within the scope of two major conferences in the field, it provides unprecedented insights into common practice among challenge participants, progress generated by competitions, open issues, as well as key winning strategies. 

A new insight with respect to common participation practice (RQ1) was that knowledge exchange is the primary participation incentive. This will most likely differ on platforms like Kaggle, in which prize money and achieving a high rank are expected to be substantially more important \cite{tauchert2020crowdsourcing}. 
To our surprise, only a small portion of participants perceived the limiting computing power as a bottleneck. Similarly surprisingly, k-fold cross-validation on the training set as well as ensembling was only performed by a minority of participants.

The competitions clearly led to substantial scientific progress according to the organizers (RQ2). Notably, however, only a small fraction of image analysis problems addressed by current competitions can be regarded as solved (App.~\ref{app:remaining_challanges}). Open research questions identified as part of this work include: \textit{(1) How can we better integrate meta information in neural network solutions?}, \textit{(2) How can we effectively leverage temporal information in biomedical video analysis?}, \textit{(3) How can we achieve generalization across devices, protocols, and sites?}, \textit{(4) How can we arrive at performance metrics that better address the biomedical domain interest?} The latter is particularly interesting in light of the fact that the reflection of metrics in the challenge design was identified as a key strategy for winning a challenge. In line with recent literature~\cite{kofler2021DICE, vaassen2020evaluation, gooding2018comparative, reinke2021common}, it implies that common efforts are focused largely on an overfitting to the current metrics rather than solving the underlying domain problem. Current initiatives are already addressing this issue~\cite{maier2022metrics}, but our results imply that challenge organizers should focus more on ensuring that the actual biomedical needs are reflected in the design of their competition.

Our work revealed particularly successful algorithm design choices (Fig.~\ref{fig:design_decisions}) and general strategies for winning a competition (Fig.~\ref{fig:winning_strategies})~(RQ3). 
%One strategy that stood out in both the survey on general recommendations and in the analysis based on linking rankings with responses is the importance of failure analysis. 
In the spirit of reporting negative results, we also included the results of the mixed model analysis despite the lack of statistical significance after correction for multiplicity of testing. Given the relatively small dataset (results from 80 competitions) compared to the number of parameters that we extracted from algorithm designs and strategies ($>100$), we hypothesize that the lack of statistical significance can largely be attributed to small sample size.

A limitation of our study could be seen in the fact that we only covered IEEE ISBI and MICCAI challenges of one specific year. Prior work, however, revealed that the competitions performed in the scope of these conferences cover the majority of all biomedical image analysis competitions~\cite{maier-hein_why_2018}. Further limitations can be regarded as general limitations when working with surveys \cite{coughlan2009survey} and include the uncertainty of self-reported data and the potential bias resulting from the preselection of categorical variables. Finally, it is not straightforward to address the heterogeneity of challenges with a single questionnaire. For example, using an in-domain similar dataset may not always be feasible due to the sparsity of public biomedical datasets. Similarly, a researcher may regard ensembling as a general key strategy but may not have had the computing power to train and optimize multiple models working with video, 3D, or 4D data. To compensate for this effect in the design of the surveys presented in \cref{sec:methods:common_practice} and \cref{sec:methods:scientific_progress}, we additionally asked winners for general recommended strategies (Fig.~\ref{fig:winning_strategies}). The discrepancy between general recommendation and feasibility is reflected in the answers. For example, most winners recommend the integration of biologists/clinicians in a team but did not do so themselves.

Despite the discussed limitations, our findings have the potential to impact a plethora of stakeholders in challenges. First, biomedical image analysis researchers and developers can ``stand on the shoulders of giants'' (the competition winners) to improve algorithm development strategies when approaching a new problem. Second, future challenge organizers can adapt their designs carefully to the open issues revealed by this work. This would include a focus on case/instance level rather than pixel/voxel level to reflect biomedical needs, metrics that reflect biomedical needs (see below), as well as dataset designs that allow for improving the capabilities of algorithms to perform well on rare classes and to generalize across domains. Given that the vast majority of participants perceived limited time and not computing power as a bottleneck, challenge timelines should be critically questioned. Finally, the wider community can benefit from the open research questions we identified (Tab.~\ref{tab:open_research_questions}).

In conclusion, we performed the first systematic analysis of biomedical image analysis competitions, which revealed a plurality of novel insights with respect to participation, organization, and winning. Our work could pave the way for (1) developers to improve algorithm development strategies when approaching new problems, and (2) the scientific community to channel its activities into open issues revealed by this work.

%%%%%%%%% REFERENCES
{\small
\bibliographystyle{ieee_fullname}
\bibliography{egbib}
}

\onecolumn

\begin{appendices}

\renewcommand{\appendixtitle}{Overview of conferences, challenges, and competitions}
\section*{\appendixtitleFull}
\label{app:overview}

\begin{longtable}{ p{0.3cm} p{0.9cm} p{2.3cm} p{2cm} p{10cm} }
\label{tab:overview:conferences} \\
\caption{Overview of conferences included in this meta-study. The conferences link to the respective conference websites. The websites were last accessed on 2022-11-11.} \\
\toprule
\# & ID & Conference & Date & Conference full name \\* \midrule
\endfirsthead
\multicolumn{5}{c}%
{{\itshape Table \thetable\ continued from previous page}} \\
\toprule
\# & ID & Conference & Date & Conference full name \\* \midrule
\endhead
1 & I & \href{https://biomedicalimaging.org/2021}{IEEE ISBI 2021} & 2021-04-13 to 2021-04-16 & 18th International Symposium on Biomedical Imaging \\
2 & M & \href{https://miccai2021.org}{MICCAI 2021} & 2021-09-27 to 2021-10-01 & 24th International Conference on Medical Image Computing and Computer Assisted Intervention \\* \bottomrule
\end{longtable}

\begin{longtable}{ p{0.3cm} p{0.9cm} p{2.3cm} p{2.8cm} p{9.2cm} }
\label{tab:overview:challenges} \\
\caption{Overview of challenges included in this meta-study. The challenge acronyms link to the respective challenge websites. The websites were last accessed on 2022-11-11.} \\
\toprule
\# & ID & Conference & Challenge acronym & Challenge full name \\* \midrule
\endfirsthead
\multicolumn{5}{c}%
{{\itshape Table \thetable\ continued from previous page}} \\
\toprule
\# & ID & Conference & Challenge acronym & Challenge full name \\* \midrule
\endhead
\bottomrule
\endfoot
\endlastfoot
1 & I.1 & IEEE ISBI 2021 & \href{http://celltrackingchallenge.net}{CTC} & 6th ISBI Cell Tracking Challenge \\
2 & I.2 & IEEE ISBI 2021 & \href{https://mitoem.grand-challenge.org}{MitoEM} & Large-scale 3D Mitochondria Instance Segmentation Challenge \\
3 & I.3 & IEEE ISBI 2021 & \href{https://endocv2021.grand-challenge.org}{EndoCV2021} & Addressing generalisability in polyp detection and segmentation challenge \\
4 & I.4 & IEEE ISBI 2021 & \href{https://riadd.grand-challenge.org}{RIADD} & Retinal Image Analysis for multi-Disease Detection Challenge \\
5 & I.5 & IEEE ISBI 2021 & \href{https://segpc-2021.grand-challenge.org/SegPC-2021}{SegPC-2021} & Segmentation of Multiple Myeloma Plasma Cells in Microscopic Images   Challenge \\
6 & I.6 & IEEE ISBI 2021 & \href{https://a-afma.grand-challenge.org}{A-AFMA} & Ultrasound Challenge: Automatic amniotic fluid measurement and analysis from ultrasound video \\
7 & M.1 & MICCAI 2021 & \href{https://kits21.kits-challenge.org}{KiTS21} & 2021 Kidney and Kidney Tumor Segmentation \\
8 & M.2 & MICCAI 2021 & \href{https://realnoisemri.grand-challenge.org}{RealNoiseMRI} & Brain MRI reconstruction challenge with realistic noise \\
9 & M.3 & MICCAI 2021 & \href{http://crossmoda-challenge.ml}{crossMoDA} & Cross-Modality Domain Adaptation for Medical Image Segmentation \\
10 & M.4 & MICCAI 2021 & \href{https://adaptor2021.github.io/}{AdaptOR 2021} & Deep Generative Model Challenge for Domain Adaptation in Surgery 2021 \\
11 & M.5 & MICCAI 2021 & \href{https://dfu-challenge.github.io/}{DFUC 2021} & Diabetic Foot Ulcer Challenge 2021 \\
12 & M.6a & MICCAI 2021 & \href{https://www.synapse.org/heisurf}{HeiSurf} & Endoscopic Vision Challenge 2021 - HeiChole Surgical Workflow Analysis and Full Scene Segmentation \\
13 & M.6b & MICCAI 2021 & \href{https://giana.grand-challenge.org/}{GIANA} & Endoscopic Vision Challenge 2021 - Gastrointestinal Image ANAlysis \\
14 & M.6c & MICCAI 2021 & \href{https://cholectriplet2021.grand-challenge.org}{CholecTriplet2021} & Endoscopic Vision Challenge 2021 - Surgical Action Triplet Recognition \\
15 & M.6d & MICCAI 2021 & \href{https://fetreg2021.grand-challenge.org}{FetReg} & Endoscopic Vision Challenge 2021 - Placental Vessel Segmentation and Registration in Fetoscopy \\
16 & M.6e & MICCAI 2021 & \href{https://www.synapse.org/PETRAW}{PETRAW} & Endoscopic Vision Challenge 2021 - PEg TRAnsfer Workflow recognition by different modalities \\
17 & M.6f & MICCAI 2021 & \href{https://www.synapse.org/simsurgskill2021}{SimSurgSkill} & Endoscopic Vision Challenge 2021 - Objective Surgical Skill Assessment in VR Simulation \\
18 & M.7 & MICCAI 2021 & \href{http://hardi.epfl.ch/static/events/2021_challenge}{DiSCo} & Diffusion-Simulated Connectivity Challenge \\
19 & M.8 & MICCAI 2021 & \href{https://flare.grand-challenge.org}{FLARE21} & Fast and Low GPU Memory Abdominal Organ Segmentation in CT \\
20 & M.9 & MICCAI 2021 & \href{https://www.med.upenn.edu/cbica/fets/miccai2021}{FeTS} & Federated Tumor Segmentation Challenge \\
21 & M.10 & MICCAI 2021 & \href{https://feta-2021.grand-challenge.org}{FeTA} & Fetal Brain Tissue Annotation and Segmentation Challenge \\
22 & M.11 & MICCAI 2021 & \href{https://www.aicrowd.com/challenges/miccai-2021-hecktor}{HECKTOR} & HEad and neCK TumOR segmentation and outcome prediction in PET/CT images \\
23 & M.12 & MICCAI 2021 & \href{https://learn2reg.grand-challenge.org}{LEARN2REG} & Learn2Reg - The Challenge (2021) \\
24 & M.13 & MICCAI 2021 & \href{http://medicalood.dkfz.de/web/2021}{MOOD} & Medical Out-of-Distribution Analysis Challenge 2021 \\
25 & M.14 & MICCAI 2021 & \href{https://imig.science/midog2021}{MIDOG} & MItosis DOmain Generalization Challenge 2021 \\
26 & M.15 & MICCAI 2021 & \href{https://www.ub.edu/mnms-2}{M\&Ms-2} & Multi-Disease, Multi-View \& Multi-Center Right Ventricular Segmentation in Cardiac MRI \\
27 & M.16 & MICCAI 2021 & \href{https://qubiq21.grand-challenge.org}{QUBIQ 2021} & Quantification of Uncertainties in Biomedical Image Quantification 2021 \\
28 & M.17 & MICCAI 2021 & \href{https://www.med.upenn.edu/cbica/brats2021}{BraTS2021} & RSNA/ASNR/MICCAI Brain Tumor Segmentation Challenge 2021 \\
29 & M.18 & MICCAI 2021 & \href{https://saras-mesad.grand-challenge.org}{SARAS-MESAD} & SARAS challenge for Multi-domain Endoscopic Surgeon Action Detection \\
30 & M.19 & MICCAI 2021 & \href{https://autoimplant2021.grand-challenge.org}{AutoImplant 2021} & Towards the Automatization of Cranial Implant Design in Cranioplasty: 2nd   MICCAI Challenge on Automatic Cranial Implant Design \\
31 & M.20 & MICCAI 2021 & \href{https://valdo.grand-challenge.org}{VALDO} & VAscular Lesions DetectiOn Challenge \\
32 & M.21 & MICCAI 2021 & \href{https://vessel-wall-segmentation.grand-challenge.org}{VWS} & Carotid Artery Vessel Wall Segmentation Challenge \\
33 & M.22 & MICCAI 2021 & \href{https://fusc.grand-challenge.org}{FU-Seg} & Foot Ulcer Segmentation Challenge 2021 \\
34 & M.23 & MICCAI 2021 & \href{https://portal.fli-iam.irisa.fr/msseg-2}{MSSEG-2} & Multiple sclerosis new lesions segmentation challenge \\
35 & M.24 & MICCAI 2021 & \href{https://paip2021.grand-challenge.org}{PAIP2021} & Perineural Invasion in Multiple Organ Cancer (Colon, Prostate, and   Pancreatobiliary tract) \\* \bottomrule
\end{longtable}

\begin{longtable}{ p{0.3cm} p{0.9cm} p{2.3cm} p{2.8cm} p{9.2cm} }
\label{tab:overview:competitions} \\
\caption{Overview of competitions included in this meta-study.} \\
\toprule
\# & ID & Conference & Challenge & Competition \\* \midrule
\endfirsthead
\multicolumn{5}{c}%
{{\itshape Table \thetable\ continued from previous page}} \\
\toprule
\# & ID & Conference & Challenge & Competition \\* \midrule
\endhead
\bottomrule
\endfoot
\endlastfoot
1 & I.1.1 & IEEE ISBI 2021 & CTC & Primary Track (evaluation across all 13 datasets) \\
2 & I.1.2 & IEEE ISBI 2021 & CTC & Secondary Track - Dataset "DIC-C2DH-HeLa" \\
3 & I.1.3 & IEEE ISBI 2021 & CTC & Secondary Track - Dataset "Fluo-C2DL-MSC" \\
4 & I.1.4 & IEEE ISBI 2021 & CTC & Secondary Track - Dataset "Fluo-C3DH-H157" \\
5 & I.1.5 & IEEE ISBI 2021 & CTC & Secondary Track - Dataset "Fluo-C3DL-MDA231" \\
6 & I.1.6 & IEEE ISBI 2021 & CTC & Secondary Track - Dataset "Fluo-N2DH-GOWT1" \\
7 & I.1.7 & IEEE ISBI 2021 & CTC & Secondary Track - Dataset "Fluo-N2DL-HeLa" \\
8 & I.1.8 & IEEE ISBI 2021 & CTC & Secondary Track - Dataset "Fluo-N3DH-CE" \\
9 & I.1.9 & IEEE ISBI 2021 & CTC & Secondary Track - Dataset "Fluo-N3DH-CHO" \\
10 & I.1.10 & IEEE ISBI 2021 & CTC & Secondary Track - Dataset "Fluo-N3DL-DRO" \\
11 & I.1.11 & IEEE ISBI 2021 & CTC & Secondary Track - Dataset "PhC-C2DH-U373" \\
12 & I.1.12 & IEEE ISBI 2021 & CTC & Secondary Track - Dataset "PhC-C2DL-PSC" \\
13 & I.1.13 & IEEE ISBI 2021 & CTC & Secondary Track - Dataset "Fluo-N2DH-SIM+" \\
14 & I.1.14 & IEEE ISBI 2021 & CTC & Secondary Track - Dataset "Fluo-N3DH-SIM+" \\
15 & I.1.15 & IEEE ISBI 2021 & CTC & Secondary Track - Dataset "BF-C2DL-HSC" \\
16 & I.1.16 & IEEE ISBI 2021 & CTC & Secondary Track - Dataset "BF-C2DL-MuSC" \\
17 & I.1.17 & IEEE ISBI 2021 & CTC & Secondary Track - Dataset "Fluo-C2DL-Huh7" \\
18 & I.1.18 & IEEE ISBI 2021 & CTC & Secondary Track - Dataset "Fluo-C3DH-A549" \\
19 & I.1.19 & IEEE ISBI 2021 & CTC & Secondary Track - Dataset "Fluo-N3DL-TRIC" \\
20 & I.1.20 & IEEE ISBI 2021 & CTC & Secondary Track - Dataset "Fluo-N3DL-TRIF" \\
21 & I.1.21 & IEEE ISBI 2021 & CTC & Secondary Track - Dataset "Fluo-C3Dh-A549-SIM" \\
22 & I.2.1 & IEEE ISBI 2021 & MitoEM & 3D Mitochondria Instance Segmentation \\
23 & I.3.1 & IEEE ISBI 2021 & EndoCV2021 & Assessing generalisability in polyp detection \\
24 & I.3.2 & IEEE ISBI 2021 & EndoCV2021 & Assessing generalisability in polyp segmentation \\
25 & I.4.1 & IEEE ISBI 2021 & RIADD & Disease Screening \\
26 & I.4.2 & IEEE ISBI 2021 & RIADD & Disease Classification \\
27 & I.5.1 & IEEE ISBI 2021 & SegPC-2021 & Segmentation of Multiple Myeloma Plasma Cells in Microscopic Images   Challenge \\
28 & I.6.1 & IEEE ISBI 2021 & A-AFMA & Detection: Automatic amniotic fluid detection from ultrasound video \\
29 & I.6.2 & IEEE ISBI 2021 & A-AFMA & Localization: Automatic amniotic fluid measurement from ultrasound video \\
30 & M.1.1 & MICCAI 2021 & KiTS21 & Segmentation of Kidney and Associated Structures \\
31 & M.2.1 & MICCAI 2021 & RealNoiseMRI & Reconstruction of motion corrupted T1 weighted MRI data \\
32 & M.2.2 & MICCAI 2021 & RealNoiseMRI & Reconstruction of motion corrupted T2 weighted MRI data \\
33 & M.3.1 & MICCAI 2021 & crossMoDA & Vestibular Schwannoma and Cochlea Segmentation \\
34 & M.4.1 & MICCAI 2021 & AdaptOR 2021 & Domain Adaptation for Landmark Detection \\
35 & M.5.1 & MICCAI 2021 & DFUC 2021 & Analysis Towards Classification of Infection \& Ischaemia of Diabetic   Foot Ulcers \\
36 & M.6a.1 & MICCAI 2021 & HeiSurf & Scene segmentation \\
37 & M.6a.2 & MICCAI 2021 & HeiSurf & Phase segmentation \\
38 & M.6a.3 & MICCAI 2021 & HeiSurf & Instrument presence \\
39 & M.6a.4 & MICCAI 2021 & HeiSurf & Action recognition \\
40 & M.6b.1 & MICCAI 2021 & GIANA & Polyp detection in colonoscopy images \\
41 & M.6b.2 & MICCAI 2021 & GIANA & Polyp segmentation in colonoscopy images \\
42 & M.6b.3 & MICCAI 2021 & GIANA & Histology prediction \\
43 & M.6c.1 & MICCAI 2021 & CholecTriplet2021 & Surgical Action Triplet Recognition \\
44 & M.6d.1 & MICCAI 2021 & FetReg & Placental semantic segmentation \\
45 & M.6d.2 & MICCAI 2021 & FetReg & Placental RGB frame registration for mosaicking \\
46 & M.6e.1 & MICCAI 2021 & PETRAW & Video-based surgical workflow recognition \\
47 & M.6e.2 & MICCAI 2021 & PETRAW & Kinematic-based surgical workflow recognition \\
48 & M.6e.3 & MICCAI 2021 & PETRAW & Segmentation-based surgical workflow recognition \\
49 & M.6e.4 & MICCAI 2021 & PETRAW & Video and kinematic-based surgical workflow recognition \\
50 & M.6e.5 & MICCAI 2021 & PETRAW & Video, kinematic and   segmentation-based surgical workflow recognition \\
51 & M.6f.1 & MICCAI 2021 & SimSurgSkill & Surgical tool/needle detection \\
52 & M.6f.2 & MICCAI 2021 & SimSurgSkill & Skill Assessment \\
53 & M.7.1 & MICCAI 2021 & DiSCo & Quantitative connectivity estimation \\
54 & M.8.1 & MICCAI 2021 & FLARE21 & Abdominal Organ Segmentation in CT Images \\
55 & M.9.1 & MICCAI 2021 & FeTS & Federated Training \\
56 & M.9.2 & MICCAI 2021 & FeTS & Federated Evaluation \\
57 & M.10.1 & MICCAI 2021 & FeTA & Fetal Brain Tissue Segmentation \\
58 & M.11.1 & MICCAI 2021 & HECKTOR & Tumor segmentation \\
59 & M.11.2 & MICCAI 2021 & HECKTOR & Radiomics \\
60 & M.11.3 & MICCAI 2021 & HECKTOR & Radiomics with ground truth contour \\
61 & M.12.1 & MICCAI 2021 & LEARN2REG & Intra-patient multimodal abdominal MRI and CT registration \\
62 & M.12.2 & MICCAI 2021 & LEARN2REG & Intra-patient large deformation lung CT registration \\
63 & M.12.3 & MICCAI 2021 & LEARN2REG & Inter-patient large scale brain MRI registration \\
64 & M.13.1 & MICCAI 2021 & MOOD & Sample-level \\
65 & M.13.2 & MICCAI 2021 & MOOD & Pixel-level \\
66 & M.14.1 & MICCAI 2021 & MIDOG & Mitotic figure detection \\
67 & M.15.1 & MICCAI 2021 & M\&Ms-2 & Segmentation of the right ventricle (RV) in cardiac MRI \\
68 & M.16.1 & MICCAI 2021 & QUBIQ 2021 & Quantifying segmentation uncertainties \\
69 & M.17.1 & MICCAI 2021 & BraTS2021 & Segmentation of glioblastoma in mpMRI scans \\
70 & M.18.1 & MICCAI 2021 & SARAS-MESAD & Multi-domain static action detection \\
71 & M.19.1 & MICCAI 2021 & AutoImplant 2021 & Cranial implant design for diverse synthetic defects on aligned skulls \\
72 & M.19.2 & MICCAI 2021 & AutoImplant 2021 & Cranial implant design for real patient defects \\
73 & M.19.3 & MICCAI 2021 & AutoImplant 2021 & Improving the model generalization ability for cranial implant design \\
74 & M.20.1 & MICCAI 2021 & VALDO & Segmentation of enlarged PVS \\
75 & M.20.2 & MICCAI 2021 & VALDO & Segmentation of cerebral microbleeds \\
76 & M.20.3 & MICCAI 2021 & VALDO & Segmentation of lacunes \\
77 & M.21.1 & MICCAI 2021 & VWS & Vessel wall segmentation \\
78 & M.22.1 & MICCAI 2021 & FU-Seg & Foot Ulcer Segmentation \\
79 & M.23.1 & MICCAI 2021 & MSSEG-2 & New MS lesions segmentation \\
80 & M.24.1 & MICCAI 2021 & PAIP2021 & Detection of perineural invasion in three organ cancers \\* \bottomrule
\end{longtable}

\renewcommand{\appendixtitle}{Profile of a competition winner}
\section*{\appendixtitleFull}
\label{app:winner_profile}

\begin{figure}[h]
  \centering
  \includegraphics[width=0.6\linewidth]
                   {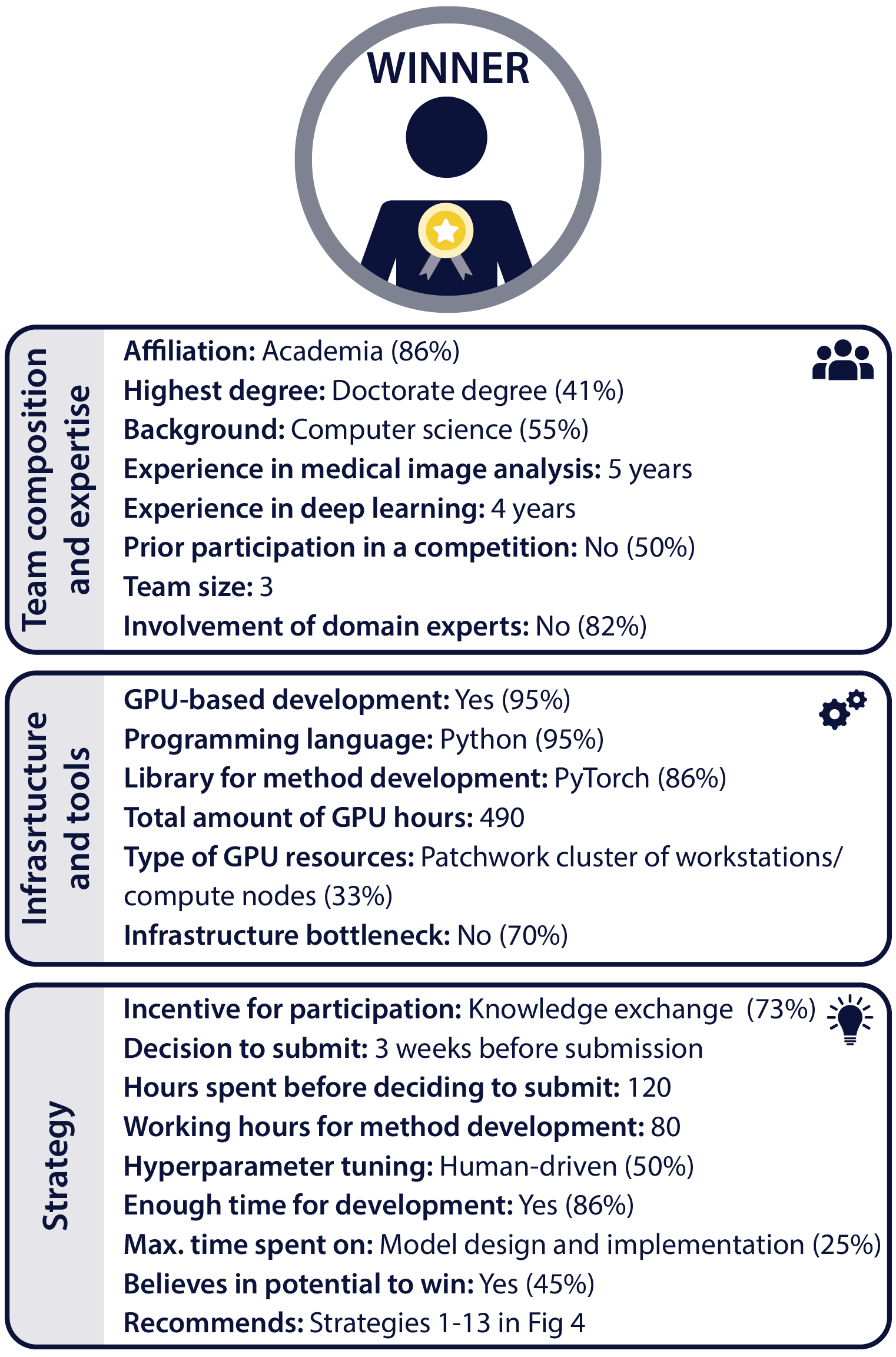}
  \caption{Profile of a competition winner. In case of categorical values, the majority vote of all participants was used (note that ``unsure'' was also an option where appropriate). In case of continuous values, the median was taken.}
  \label{fig:winner_profile}
\end{figure}

\renewcommand{\appendixtitle}{Remaining challenges}
\section*{\appendixtitleFull}
\label{app:remaining_challanges}

The amount of tasks that were ``not solved at all''/``completely solved'' differed substantially across medical disciplines ranging from 0\%/33\% for pathology, 0\%/8\% for radiology, to 16\%/12\% for surgery, thus suggesting that specifically surgical video analysis is an open grand challenge. We have further obtained permission from the challenge organizers to de-anonymize their answers in order to explicitly list the gaps in the literature for tasks that have not been solved (Tab.~\ref{tab:remaining_challanges}). Some overarching open research questions extracted from our work are provided in Tab.~\ref{tab:open_research_questions}.

\begin{longtable}{ p{6.0cm} p{1.5cm} p{9.0cm} }
\label{tab:remaining_challanges} \\
\caption{Remaining challenges related to unsolved biomedical imaging tasks according to challenge organizers. The task IDs given in brackets refer to App.~\ref{app:overview}.} \\
\toprule
Task & Solved? & Core remaining challenges \\* \midrule
\endfirsthead
\multicolumn{3}{c}%
{{\itshape Table \thetable\ continued from previous page}} \\
\toprule
Task & Solved? & Core remaining challenges \\* \midrule
\endhead
ISBI / MitoEM (I.2.1): 3D mitochondria instance segmentation & No; \newline substantial progress & Segmentation of a specific type of mitochondria (mitochondria-on-a-string); imaging artifacts such as knife marks and lighting variations. \\
ISBI / SegPC-2021 (I.5.1): Segmentation of multiple myeloma plasma cells in microscopic images & No; minor progress & Cytoplasm boundaries, which provide a poor contrast relative to background. \\
ISBI / A-AFMA (I.6.1): Automatic amniotic fluid detection from ultrasound video & No; first baseline & Lack of use of temporal context; lack of  incorporation of clinical context knowledge; poor imaging conditions (e.g., artifacts). \\
ISBI / A-AFMA (I.6.2): Automatic amniotic fluid measurement from ultrasound video & No; minor progress & Ultrasound image artifacts, e.g., acoustic shadow, unclear boundaries. \\
MICCAI / KiTS21 (M.1.1): Segmentation of kidney and associated structures & No; \newline substantial progress & Small cysts and tumors. \\
MICCAI / RealNoiseMRI (M.2.1): Reconstruction of motion corrupted T1 weighted MRI data & No; minor progress & The challenge was focused on providing realistically degraded brain MRI to evaluate reconstruction algorithms on, and it showed that current algorithms still cannot handle this. \\
MICCAI / RealNoiseMRI (M.2.2): Reconstruction of   motion corrupted T2 weighted MRI data & No;   minor progress & The   challenge was focused on providing realistically degraded brain MRI to   evaluate reconstruction algorithms on, and it showed that current algorithms   still cannot handle this. \\
MICCAI / AdaptOR 2021 (M.4.1): Domain Adaptation   for landmark detection & Not   at all & Varying   light conditions and non-standardized views. \\
MICCAI / DFUC 2021 (M.5.1): Analysis towards classification of infection \& ischaemia of diabetic foot ulcers & No; \newline substantial progress & Data imbalance (the majority of the images are of type infection and control). \\
MICCAI / EndoVis / HeiSurf (M.6a.1): Scene segmentation & No; \newline substantial progress & Smoke   and motion blur, high brightness/darkness. \\
MICCAI / EndoVis / HeiSurf (M.6a.2): Phase segmentation & No; \newline substantial progress & Complex   surgeries, out-of-body frames. \\
MICCAI / EndoVis / HeiSurf (M.6a.3): \newline Instrument presence & No; \newline substantial progress & Occlusions,   and motion blur. \\
MICCAI / EndoVis / HeiSurf (M.6a.4): Action recognition & No;   minor progress & Poor   performance on most classes: only one of four actions (hold) is reliably   detected. \\
MICCAI / EndoVis / GIANA (M.6b.3): Histology prediction & No; \newline substantial progress & Poor   performance on the rare class (adenomatous). \\
MICCAI / EndoVis / CholecTriplet (M.6c.1):   Surgical action recognition & No;    minor progress & Non-intuitive   data augmentation; poor performance on rare triplets and difficult imaging   conditions. \\
MICCAI / EndoVis / FetReg (M.6d.1): Placental semantic   segmentation & No;   minor progress & High   variability across different centers/devices; poor visibility, varying   illumination, artifacts, occlusions. \\
MICCAI / EndoVis / FetReg (M.6d.2): Placental RGB frame   registration for mosaicking & Not   at all & Poor   vessel visibility, artifacts, texture paucity, occlusions, non-planar views,   non-rigid deformations (i.e., maternal breathing). \\
MICCAI / EndoVis / PETRAW (M.6e.1): Video-based surgical   workflow recognition & No;   minor progress & Non-standardized   workflows; the data set is highly unbalanced with respect to the action verb   (some prevalences \textless 5\%). \\
MICCAI / EndoVis / PETRAW (M.6e.2): Kinematic-based   surgical workflow recognition & No;   minor progress & Non-standardized   workflows; the data set is highly unbalanced with respect to the action verb   (some prevalences \textless 5\%). \\
MICCAI / EndoVis / PETRAW (M.6e.3): Segmentation-based   surgical workflow recognition & Not   at all & Non-standardized   workflows; the data set is highly unbalanced with respect to the action verb   (some prevalences \textless 5\%). \\
MICCAI / EndoVis / PETRAW (M.6e.4): Video and   kinematic-based surgical workflow recognition & No;   minor progress & The   data set is highly unbalanced with respect to the action verb (some   prevalences \textless 5\%). \\
MICCAI / EndoVis / PETRAW (M.6e.5): Video, kinematic and   segmentation-based surgical workflow recognition & No;   minor progress & Non-standardized   workflows; the data set is highly unbalanced with respect to the action verb   (some prevalences \textless 5\%). \\
MICCAI / EndoVis / SimSurgSkill (M.6f.1): Skill Assessment & No; minor progress & Change   of background. \\
MICCAI / EndoVis / SimSurgSkill (M.6f.2): Surgical   tool/needle detection & No; \newline substantial progress & Domain   shifts (i.e., different background but same tools). \\
MICCAI / DiSCo (M.7.1): Quantitative   connectivity estimation & No; \newline substantial progress & Estimation   of connectivity in more complex microstructural environments; coping with   various MRI noise sources and artifacts. \\
MICCAI / FLARE21 (M.8.1): Abdominal organ   segmentation in CT images & No; \newline substantial progress & The   top algorithms tend to fail in the regions with low contrast, pathological   changes, and fuzzy boundaries. \\
MICCAI / FeTS (M.9.2): Federated evaluation & No; minor progress & Reference   segmentations with empty enhancing tumor regions lead to extreme metric values. Analysis of failure cases is limited by the federated setup. \\
MICCAI / FeTA (M.10.1): Fetal brain tissue   segmentation & No; \newline substantial progress & Consistent   performance across all five classes. \\
MICCAI / HECKTOR (M.11.1): \newline Tumor segmentation & No; \newline substantial progress & Corner   cases, such as large metastatic lymph nodes incorrectly segmented as a primary tumor. \\
MICCAI / HECKTOR (M.11.2): \newline Radiomics & No; \newline substantial progress & Stratifications   of patient populations in terms of treatment and imaging protocols, HPV status, and age are needed. \\
MICCAI / HECKTOR (M.11.3): \newline Radiomics with ground truth contours & No; \newline substantial progress & Stratifications of patient populations in terms of treatment and imaging protocols, HPV   status, and age are needed. \\
MICCAI / LEARN2REG (M.12.1): Intra-patient   multimodal abdominal MRI and CT registration & No; \newline substantial progress & Large   deformations, large variations in voxel-size (before pre-processing), and   contrast. \\
MICCAI / MOOD (M.13.1): Pixel-level & No;   minor progress & Medically   relevant but less obvious anomalies. \\
MICCAI / MOOD (M.13.2): Sample-level & No;   minor progress & Medically   relevant but less obvious anomalies. \\
MICCAI / SARAS-MESAD (M.18.1): Multi-domain   static action detection & Not   at all & Domain   shifts (i.e., large shifts in the characteristics of scenes, such as change   in lighting conditions, properties of the tools, change in endoscopic   camera). \\
MICCAI / AutoImplant 2021 (M.19.1): Cranial   implant design for diverse synthetic defects on aligned skulls & No; \newline substantial progress & Complex   shapes in the "frontal" group of implant cases. \\
MICCAI / AutoImplant 2021 (M.19.2): Cranial   implant design for real patient defects & No;   minor progress & Big   cranial defects. The given training data was strictly shape completion, but   the predicted implants were evaluated against real cranial implants. This led   to a mismatch in the shape of the predicted implant designs. \\
MICCAI / Where is VALDO (M.20.3): Segmentation   of lacunes & No;   minor progress & Dataset   representativeness, e.g., sparsity of small elements of interest with   potentially large variability in shape and intensity signature. \\
MICCAI / VWS (M.21.1): Vessel wall segmentation & No; \newline substantial progress & Vessel   wall segmentation at the carotid bifurcation, in the presence of flow   artifacts or low SNR; complex plaque.\\* \bottomrule
\end{longtable}

\renewcommand{\appendixtitle}{Open research questions}
\section*{\appendixtitleFull}
\label{app:open_research_questions}

\begin{longtable}{ p{17cm} }
\label{tab:open_research_questions} \\
\caption{Overarching general open research questions extracted from this work.} \\
\toprule
Research question \\* \midrule
\endfirsthead
\multicolumn{1}{c}%
{{\itshape Table \thetable\ continued from previous page}} \\
\toprule
Research question \\* \midrule
\endhead
How can we better integrate clinical context data in neural network solutions? \\
How can we effectively leverage temporal information in endoscopic video analysis? \\
How to do data augmentation in endoscopic video analysis? \\
How can we achieve generalization across biomedical devices, protocols, and caregivers involved in image acquisition? \\
How can we arrive at performance metrics that better address the biomedical domain interest? \\* \bottomrule
\end{longtable}

\renewcommand{\appendixtitle}{Full affiliation list}
\section*{\appendixtitleFull}
\label{app:affiliation_list}

% Please don't touch anything here - this content was generated automatically.
\textsuperscript{1}Division of Intelligent Medical Systems, German Cancer Research Center (DKFZ), Heidelberg, Germany;
\textsuperscript{2}Helmholtz Imaging, German Cancer Research Center (DKFZ), Heidelberg, Germany;
\textsuperscript{3}Faculty of Mathematics and Computer Science, Heidelberg University, Heidelberg, Germany;
\textsuperscript{4}Division of Biostatistics, German Cancer Research Center (DKFZ), Heidelberg, Germany;
\textsuperscript{5}Division of Medical Image Computing, German Cancer Research Center (DKFZ), Heidelberg, Germany;
\textsuperscript{6}School of Computing, Faculty of Engineering and Physical Sciences, University of Leeds, Leeds, UK;
\textsuperscript{7}Institute of Informatics, School of Management, HES-SO Valais-Wallis University of Applied Sciences and Arts Western Switzerland, Sierre, Switzerland;
\textsuperscript{8}Department of Nuclear Medicine and Molecular Imaging, Lausanne University Hospital, Lausanne, Switzerland;
\textsuperscript{9}Technische Hochschule Ingolstadt, Ingolstadt, Germany;
\textsuperscript{10}Center for Artificial Intelligence and Data Science for Integrated Diagnostics (AI\textsuperscript{2}D) and Center for Biomedical Image Computing and Analytics (CBICA), University of Pennsylvania, Philadelphia, PA, USA;
\textsuperscript{11}Department of Pathology and Laboratory Medicine, Perelman School of Medicine, University of Pennsylvania, Philadelphia, PA, USA;
\textsuperscript{12}Department of Radiology, Perelman School of Medicine, University of Pennsylvania, Philadelphia, PA, USA;
\textsuperscript{13}Department of Radiology, University of Washington, Seattle, WA, USA;
\textsuperscript{14}Wellcome/EPSRC Centre for Interventional and Surgical Sciences (WEISS) and Department of Computer Science, University College London, London, UK;
\textsuperscript{15}Universitat Autònoma de Barcelona \& Computer Vision Center, Barcelona, Spain;
\textsuperscript{16}Division of Translational Surgical Oncology, National Center for Tumor Diseases (NCT/UCC) Dresden, Dresden, Germany;
\textsuperscript{17}Department of Advanced Robotics, Istituto Italiano di Tecnologia, Italy and Department of Electronics, Information and Bioengineering, Politecnico di Milano, Milan, Italy;
\textsuperscript{18}IT University of Copenhagen, Copenhagen, Denmark;
\textsuperscript{19}Department of General, Visceral and Transplantation Surgery, Heidelberg University Hospital, Heidelberg, Germany;
\textsuperscript{20}Biomedical Imaging Group Rotterdam, Department of Radiology and Nuclear Medicine, Erasmus MC, Rotterdam, The Netherlands;
\textsuperscript{21}Department of Computer Science, University of Copenhagen, Copenhagen, Denmark;
\textsuperscript{22}Institute of Information Systems, University of Applied Sciences Western Switzerland (HES-SO), Sierre, Switzerland;
\textsuperscript{23}Harvard Medical School, Brigham and Women’s Hospital, Boston, MA, USA;
\textsuperscript{24}School of Biomedical Engineering and Imaging Sciences, King's College London, London, UK;
\textsuperscript{25}Institute for Artificial Intelligence in Medicine (IKIM), University Hospital Essen (AöR), Essen, Germany;
\textsuperscript{26}University of Nebraska Medical Center, Omaha, NE, USA;
\textsuperscript{27}Department of Internal Medicine III, Heidelberg University Hospital, Heidelberg, Germany;
\textsuperscript{28}Neurobiology Research Unit, Copenhagen University Hospital, Rigshospitalet, Copenhagen, Denmark;
\textsuperscript{29}Arab Academy of Science and Technology, Cairo, Egypt;
\textsuperscript{30}CIBM Center for Biomedical Imaging, Lausanne, Switzerland;
\textsuperscript{31}Radiology Department, Centre Hospitalier Universitaire Vaudois (CHUV) and University of Lausanne (UNIL), Lausanne, Switzerland;
\textsuperscript{32}Signal Processing Laboratory (LTS5), École Polytechnique Fédérale de Lausanne (EPFL), Lausanne, Switzerland;
\textsuperscript{33}National Center for Tumor Diseases (NCT), Heidelberg, Germany;
\textsuperscript{34}SBILab, Department of ECE, IIIT-Delhi, Delhi, India;
\textsuperscript{35}University of Lübeck, Lübeck, Germany;
\textsuperscript{36}Mechanical Engineering, School of Engineering, The University of Tokyo, Tokyo, Japan;
\textsuperscript{37}University of Minnesota, Department of Computer Science \& Engineering, Minneapolis, MN, USA;
\textsuperscript{38}Diagnostic Image Analysis Group, Radboud University Medical Center, Nijmegen, The Netherlands;
\textsuperscript{39}Fraunhofer MEVIS, Lübeck, Germany;
\textsuperscript{40}Univ Rennes, INSERM, LTSI - UMR 1099, F35000, Rennes, France;
\textsuperscript{41}Brno University of Technology, Brno, Czech Republic;
\textsuperscript{42}Centre for Biomedical Image Analysis, Masaryk University, Brno, Czech Republic;
\textsuperscript{43}Department of Quantitative Biomedicine, University of Zurich, Zurich, Switzerland;
\textsuperscript{44}Laboratory Medicine and Pathobiology, University of Toronto, Toronto, Canada;
\textsuperscript{45}Departament de Matemàtiques \& Informàtica, Universitat de Barcelona, Barcelona, Spain;
\textsuperscript{46}Biomedical Image Analysis \& Machine Learning, Department of Quantitative Biomedicine, University of Zurich, Zurich, Switzerland;
\textsuperscript{47}Department of Engineering Science, University of Oxford, Oxford, UK;
\textsuperscript{48}Institute of Informatics, University of Applied Sciences Western Switzerland (HES-SO), Sierre, Switzerland;
\textsuperscript{49}ICube, University of Strasbourg, CNRS, Strasbourg, France;
\textsuperscript{50}IHU Strasbourg, Strasbourg, France;
\textsuperscript{51}Center For Artificial Intelligence And Data Science For Integrated Diagnostics (AI\textsuperscript{2}D) and Center for Biomedical Image Computing and Analytics (CBICA), University of Pennsylvania, Philadelphia, PA, USA;
\textsuperscript{52}Department of Informatics, Technical University of Munich, Munich, Germany;
\textsuperscript{53}Center for MR Research, University Children’s Hospital Zurich, University of Zurich, Zurich, Switzerland;
\textsuperscript{54}Neuroscience Center Zurich, University of Zurich, Zurich, Switzerland;
\textsuperscript{55}Visual Artificial Intelligence Laboratory (VAIL), Oxford Brookes University, Oxford, UK;
\textsuperscript{56}Centre for Tactile Internet with Human-in-the-Loop (CeTI), TU Dresden, Dresden, Germany;
\textsuperscript{57}MRC Unit for Lifelong Health and Ageing, University College London, London, UK;
\textsuperscript{58}Centre for Medical Image Computing, University College London, London, UK;
\textsuperscript{59}School of Biomedical Engineering \& Imaging Sciences, King's College London, London, UK;
\textsuperscript{60}Dementia Research Centre, University College London, London, UK;
\textsuperscript{61}Computer Science, Boston College, Boston, USA;
\textsuperscript{62}Department of Computing and Mathematics, Manchester Metropolitan University, Manchester, UK;
\textsuperscript{63}Medical Faculty, Heidelberg University, Heidelberg, Germany;
\textsuperscript{64}Intuitive Surgical, Inc., Sunnyvale, CA, USA;
\textsuperscript{65}A.I. Virtanen Institute for Molecular Sciences, University of Eastern Finland, Kuopio, Finland;
\textsuperscript{66}Department of Neuroscience and Biomedical Engineering, Aalto University School of Science, Espoo, Finland;
\textsuperscript{67}University of Aberdeen, Aberdeen, UK;
\textsuperscript{68}Department of Computer Science, University of Applied Sciences and Arts Dortmund, Dortmund, Germany;
\textsuperscript{69}Institute for Medical Informatics, Biometry and Epidemiology (IMIBE), University Hospital Essen, Essen, Germany;
\textsuperscript{70}Institute for Artificial Intelligence in Medicine (IKIM), University Hospital Essen, Essen, Germany;
\textsuperscript{71}School of Computing, Korea Advanced Institute of Science and Technology (KAIST), Daejeon, Republic of Korea;
\textsuperscript{72}Department of Electrical Engineering and Computer Science, Massachusetts Institute of Technology, Cambridge, MA, USA;
\textsuperscript{73}The Chinese University of Hong Kong, Hong Kong;
\textsuperscript{74}Department of Computer Science, Technical University of Munich, Munich, Germany;
\textsuperscript{75}The University of Texas MD Anderson Cancer Center, Houston, TX, USA;
\textsuperscript{76}Nepal Applied Mathematics and Informatics Institute for Research (NAAMII), Lalitpur, Nepal;
\textsuperscript{77}Universidad Pompeu Fabra, Barcelona, Spain;
\textsuperscript{78}University of Adelaide, Australia, Australia;
\textsuperscript{79}XLAB d.o.o., Ljubljana, Slovenia;
\textsuperscript{80}Touch Surgery, Medtronic, London, UK;
\textsuperscript{81}CJ AI Center, Seoul, Republic of Korea;
\textsuperscript{82}Tissue Image Analytics Centre, Department of Computer Science, University of Warwick, Coventry, UK;
\textsuperscript{83}Hankuk University of Foreign Studies, Yongin, Republic of Korea;
\textsuperscript{84}Massachusetts General Hospital, Boston, MA, USA;
\textsuperscript{85}Harvard Medical School, Boston, MA, USA;
\textsuperscript{86}Helmholtz AI, Helmholtz Zentrum München, Munich, Germany;
\textsuperscript{87}Department of Informatics, Technical University Munich, Munich, Germany;
\textsuperscript{88}TranslaTUM - Central Institute for Translational Cancer Research, Technical University of Munich, Munich, Germany;
\textsuperscript{89}Department of Diagnostic and Interventional Neuroradiology, School of Medicine, Klinikum rechts der Isar, Technical University of Munich, Munich, Germany;
\textsuperscript{90}Muroran Institute of Technology, Hokkaido, Japan;
\textsuperscript{91}UMC Utrecht, Utrecht, The Netherlands;
\textsuperscript{92}University of Science and Technology of China, Hefei, China;
\textsuperscript{93}Department of Bio and Brain Engineering, Korea Advanced Institute of Science and Technology (KAIST), Daejeon, Republic of Korea;
\textsuperscript{94}2Ai, School of Technology, IPCA, Barcelos, Portugal;
\textsuperscript{95}Algoritmi Center, School of Engineering, University of Minho, Guimarães, Portugal;
\textsuperscript{96}Life and Health Sciences Research Institute, School of Medicine, University of Minho, Braga, Portugal;
\textsuperscript{97}Department of Mechanical Engineering, Massachusetts Institute of Technology, Cambridge, MA, USA;
\textsuperscript{98}Sano Centre for Computational Medicine, Cracow, Poland;
\textsuperscript{99}Informatics Institute, University of Amsterdam, Amsterdam, The Netherlands;
\textsuperscript{100}Department of Biomedical Engineering and Physics, Amsterdam University Medical Center, University of Amsterdam, Amsterdam, The Netherlands;
\textsuperscript{101}LRE, EPITA, Paris, France;
\textsuperscript{102}Artificial Intelligence and Robotics Institute, Korea Institute of Science and Technology, Seoul, Republic of Korea;
\textsuperscript{103}Mohamed Bin Zayed University of Artificial Intelligence, Abu Dhabi, UAE;
\textsuperscript{104}Harbin Institute of Technology, Shenzhen, China;
\textsuperscript{105}University of Ljubljana, Faculty of Computer and Information Science, Ljubljana, Slovenia;
\textsuperscript{106}ISEP, Paris, France;
\textsuperscript{107}Department of Computer Science at School of Informatics, Xiamen University, Xiamen, China;
\textsuperscript{108}College of Computer Science, Sichuan University, Chengdu, China;
\textsuperscript{109}Department of Neuroradiology, Technical University of Munich, Munich, Germany;
\textsuperscript{110}AGH UST, Department of Measurement and Electronics, Kraków, Poland;
\textsuperscript{111}University of Applied Sciences and Arts Western Switzerland (HES-SO), Sierre, Switzerland;
\textsuperscript{112}Data Science and Learning Division, Argonne National Laboratory, Lemont, IL, USA;
\textsuperscript{113}University of Chicago, Chicago, IL, USA;
\textsuperscript{114}Shaanxi Normal University, Xi'an, China;
\textsuperscript{115}AI Lab, Tencent, Shenzhen, China;
\textsuperscript{116}Pattern Analysis and Learning Group, Department of Radiation Oncology, Heidelberg University Hospital, Heidelberg, Germany;
\textsuperscript{117}Interactive Machine Learning Group, German Cancer Research Center (DKFZ), Heidelberg, Germany

\renewcommand{\appendixtitle}{Acknowledgments}
\section*{\appendixtitleFull}
\label{app:acknowledgments}

Part of this work was funded by Helmholtz Imaging, a platform of the Helmholtz Information \& Data Science Incubator. S. Bakas was supported by the National Institutes of Health (NIH) award number NCI/ITCR:U01CA242871. J. Bernal acknowledges funding from MCIN/AEI/10.13039/501100011033, Grant Number: PID2020-120311RB-I00. M. de Bruijne was supported by the Netherlands Organisation for Scientific Research (NWO) project VI.C.182.042. R. Dorent received funding from the Engineering and Physical Sciences Research Council (EPSRC) [NS/A000049/1, NS/A000050/1], MRC (MC/PC/180520), Wellcome Trust [203145Z/16/Z, 203148/Z/16/Z, WT106882], and the National Institutes of Health via P41EB028741. A. Gupta acknowledges funding from Grant No. EMR/2016/006183 by SERB, Department of Science \& Technology, Govt. of India. N. Heller was partially supported by the National Institutes of Health under award number R01CA225435. M. Kozubek was supported by the Czech Ministry of Education, Youth and Sports (Project LM2023050). B. Menze acknowledges funding from the Helmut Horten Stiftung. V. Singh Bawa was funded by the European Union's Horizon 2020 research and innovation programme under grant agreement No. 779813. K. van Wijnen received funding from the Dutch Technology Foundation STW, project P15-26. L. Bloch and R. Brüngel were partially funded by a PhD grant from the University of Applied Sciences and Arts Dortmund (FH Dortmund), Dortmund, Germany. I. Ezhov acknowledges funding from the TUM International Graduate School of Science and Engineering (IGSSE). A. Galdran was supported by a Marie Sklodowska-Curie Fellowship (No 892297). Á. García Faura, T. Martinčič and D. Štepec were partially supported by the European Commission through Horizon 2020 and Horizon Europe research and innovation programs under grants 826121 (iPC), 101057294 (AIDEAS) and 101073821 (SUNRISE). F. Kofler was supported through the SFB 824, subproject B12, supported by Anna Valentina Lioba Eleonora Claire Javid Mamasani, and by Deutsche Forschungsgemeinschaft (DFG) through TUM International Graduate School of Science and Engineering (IGSSE), GSC 81. P. Morais was funded by national funds, through the FCT (Fundação para a Ciência e a Tecnologia) and FCT/MCTES in the scope of the project UIDB/05549/2020 and CEECINST/00039/2021. S. Płotka was funded by the European Union’s Horizon 2020 research and innovation programme under grant agreement no. 857533 (Sano) and the International Research Agendas programme of the Foundation for Polish Science, co-financed by the European Union under the European Regional Development Fund. J. L. Vilaça was partially supported by the project NORTE-01-0145-FEDER-000045 funded by the Northern Portugal Regional Operational Programme (NORTE 2020), under the Portugal 2020 Partnership Agreement, through the European Regional Development Fund (FEDER). K. A. Wahid was supported by the National Institutes of Health (NIH) grant no. 5F31DE031502-02. H. Torres was funded by FCT, Portugal and the European Social Found, European Union, through the ``Programa Operacional Capital Humano'' (POCH) in the scope of the PhD grants SFRH/BD/136670. B. Oliveira was funded by FCT, Portugal and the European Social Found, European Union, through the ``Programa Operacional Capital Humano'' (POCH) in the scope of the PhD grant SFRH/BD/136721/2018. N. Padoy and his team were supported by French state funds managed within the Investissements d'Avenir program by BPI France (project CONDOR) and by the ANR (references ANR-11-LABX-0004, ANR-16-CE33-0009 and ANR-10-IAHU-02). I. Jang was supported by Hankuk University of Foreign Studies Research Fund of 2023. A. Casella was supported by Giancarlo Ferrigno, Elena De Momi (Politecnico di Milano, Milan, Italy) and Leonardo S. Mattos (Istituto Italiano di Tecnologia, Genoa, Italy). M. Zenk was partially funded by the Helmholtz Association within the project ``Trustworthy Federated Data Analytics'' (TFDA) (funding number ZT-I-OO1~4).

We thank Susanne Steger (Data Protection Office, DKFZ) for the data protection supervision and Anke Trotter (IT Core Facility, DKFZ) for the hosting of the surveys.

We further thank Tony C. W. Mok (The Hong Kong University of Science and Technology, Hong Kong), Vajira Thambawita (Simula Research Laboratory, Norway), and Amirreza Mahbod (Medical University of Vienna, Austria) for taking the winner survey.

We thank Sergio Escalera Guerrero (Universitat de Barcelona and Computer Vision Center, Spain), and Karim Lekadir (Universitat de Barcelona, Spain) for supervision of challenge participants.

\end{appendices}

\end{document}